\documentclass[10pt,twocolumn,letterpaper]{article}

\usepackage{cvpr}
\usepackage{times}
\usepackage{epsfig}
\usepackage{graphicx}
\usepackage{amsmath}
\usepackage{amssymb}
\usepackage{tabularx}
\usepackage{multirow,array}
\usepackage{booktabs}
\usepackage{bm}
\usepackage{url}

\usepackage[breaklinks=true,bookmarks=false]{hyperref}

\cvprfinalcopy 

\def\cvprPaperID{4463} 

\ifcvprfinal\fi
\begin{document}

\title{VITAMIN-E: VIsual Tracking And MappINg \\with Extremely Dense Feature Points}

\author{Masashi Yokozuka, Shuji Oishi, Thompson Simon, Atsuhiko Banno\\
Robot Innovation Research Center, \\
National Institute of Advanced Industrial Science and Technology (AIST), Japan\\
{\tt\small \{yokotsuka-masashi, shuji.oishi, simon.thompson, atsuhiko.banno\}@aist.go.jp}
}

\maketitle

\begin{abstract}
In this paper, we propose a novel indirect monocular SLAM algorithm called ``VITAMIN-E,'' which is highly accurate and robust as a result of tracking extremely dense feature points.
Typical indirect methods have difficulty in reconstructing dense geometry because of their careful feature point selection for accurate matching.
Unlike conventional methods, the proposed method processes an enormous number of feature points by tracking the local extrema of curvature informed by dominant flow estimation.
Because this may lead to high computational cost during bundle adjustment, we propose a novel optimization technique, the "subspace Gauss--Newton method", that significantly improves the computational efficiency of bundle adjustment by partially updating the variables.
We concurrently generate meshes from the reconstructed points and merge them for an entire 3D model.
The experimental results on the SLAM benchmark dataset EuRoC demonstrated that the proposed method outperformed state-of-the-art SLAM methods, such as DSO, ORB-SLAM, and LSD-SLAM, both in terms of accuracy and robustness in trajectory estimation.
The proposed method simultaneously generated significantly detailed 3D geometry from the dense feature points in real time using only a CPU.
\end{abstract}


\begin{figure}[!t]
    \centering
    \includegraphics[width=0.95\linewidth]{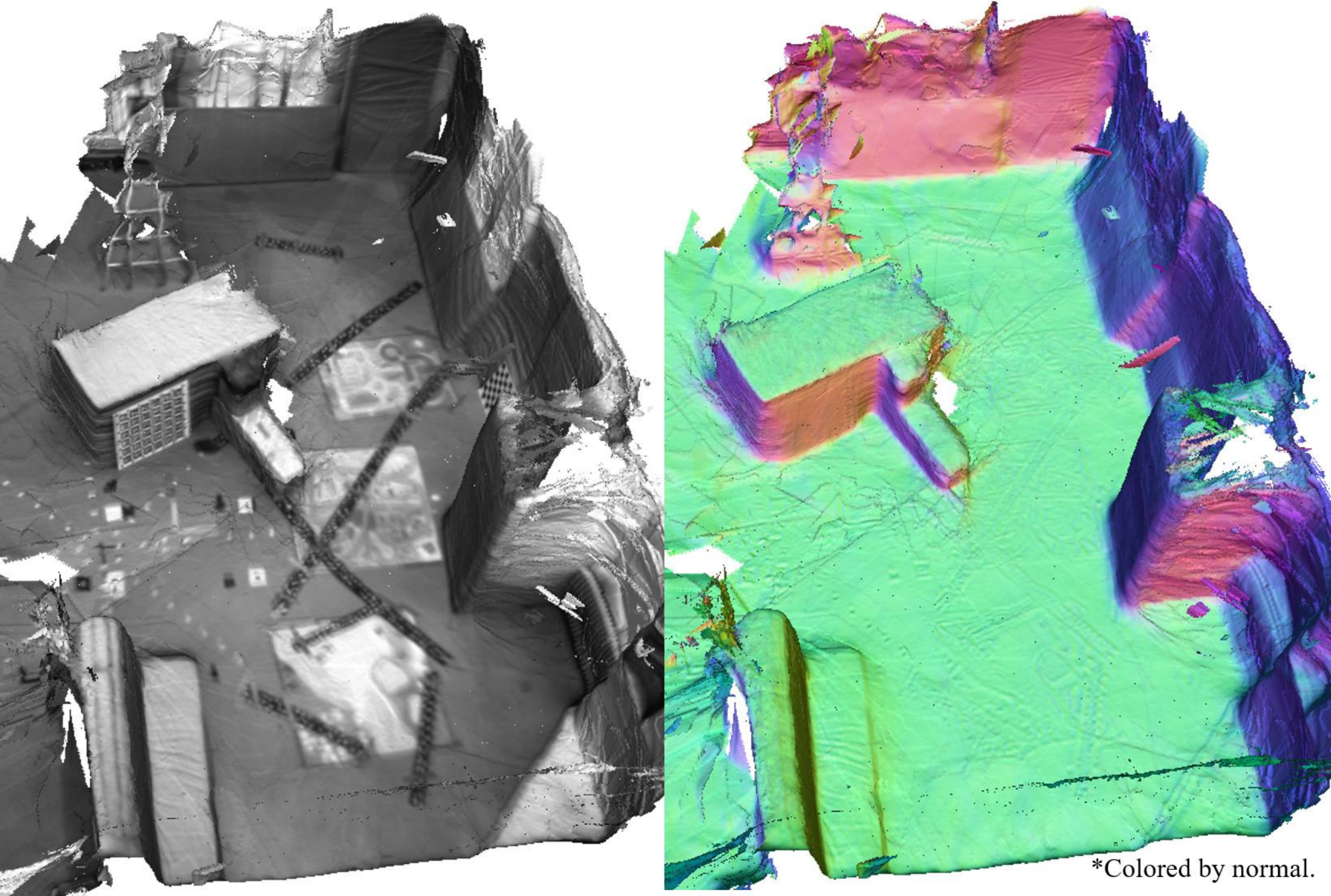}
    \caption{
        Dense geometry reconstruction with VITAMIN-E on EuRoC V101.
        (\url{https://youtu.be/yfKccCmmMsM})
    }
    \label{fig:FirstFig}
\end{figure}

\begin{figure*}[!t]
    \centering
    \includegraphics[width=0.95\linewidth]{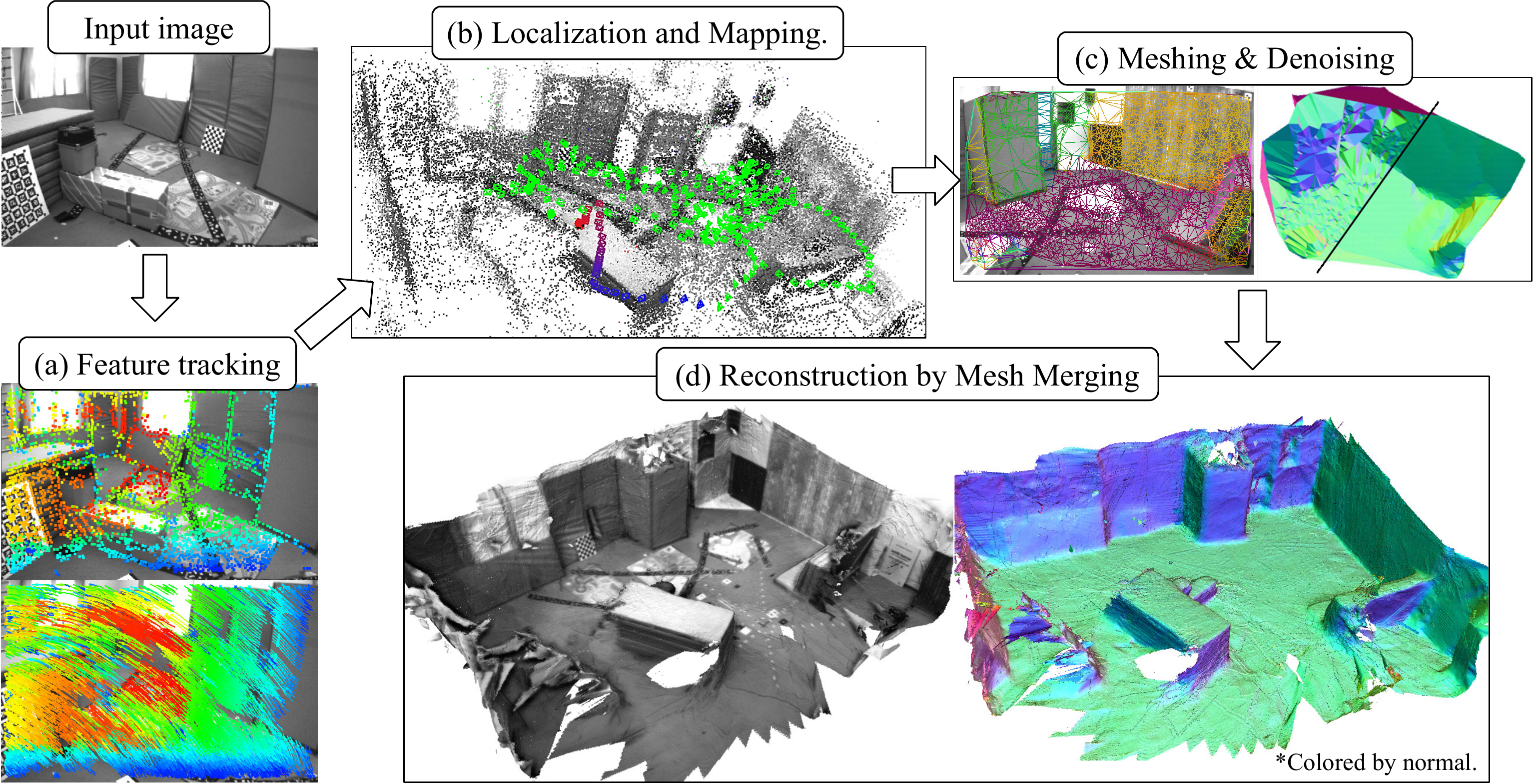}
    \caption{
        Overview of the proposed monocular SLAM.
    }
    \label{fig:MainArt}
\end{figure*}

\section{Introduction}

Simultaneous localization and mapping (SLAM) is a key technology for applications such as autonomous systems and augmented reality.
Whereas LiDAR-based SLAM\cite{LidarSLAM1, LidarSLAM2, LidarSLAM3} is well established and widely used in autonomous vehicles, visual SLAM with a monocular camera does not provide sufficient accuracy and robustness, particularly regarding dense map reconstruction, to replace LiDAR-based SLAM.
Although some visual SLAM algorithms that use stereo cameras\cite{Stereo1, Stereo2}, RGB-D cameras\cite{RGBD1, RGBD2, RGBD3}, and inertial sensors\cite{VIO1, VIO2, VIO3, VIO4, VIO5} have achieved high performance, these methods are based on pure monocular SLAM; hence, improving monocular SLAM is important.

Monocular SLAM methods can be classified into two types: direct methods and indirect methods.

{\bf Direct methods}: Direct methods estimate camera poses and reconstruct the scene by minimizing the photometric error defined as a sum of the intensity difference between each pixel in the latest image and the reprojection of the color / monochrome 3D map.
Direct methods, such as LSD-SLAM\cite{LSD-SLAM}, SVO\cite{SVO}, and DSO\cite{DSO}, process almost all pixels in incoming images.
They do not require exact pixel correspondences among multiple views unlike indirect methods, which leads to denser map reconstruction.
However, direct methods are susceptible to image noise, luminance fluctuation, and lens aberration because they directly use pixel intensities.
To overcome this drawback, Bergmann et al.\cite{PhotoCalib} proposed a normalization method against luminance fluctuation and a calibration method for lens aberration.
As another approach, Zhang et al.\cite{AutoExpo} proposed an auto-exposure method that is suitable for direct methods.

{\bf Indirect methods}: Indirect methods minimize the geometric error between observed 2D feature points and reprojections of the corresponding 3D points.
As a result of the use of feature descriptors, indirect methods such as PTAM\cite{PTAM} and ORB-SLAM\cite{ORB-SLAM} are robust against brightness changes and image noise.
Additionally, indirect methods explicitly establish feature point correspondences; hence, outliers are easily removed using RANSAC\cite{RANSAC} or M-estimation\cite{C}.
This characteristic, however, can be a drawback: Indirect methods carefully select stable feature points, thus the reconstructed 3D map tends to be sparse and does not provide detailed geometry.
Densification methods, such as PMVS\cite{PMVS} and the extension L-PMVS\cite{L-PMVS}, might be useful for obtaining dense geometry; however, they are offline methods and not applicable in real time.

In this paper, we propose the novel VIsual Tracking And MappINg with Extremely dense feature points, ``VITAMIN-E,'' which is highly precise, robust, and dense because of the tracking of a large number of feature points.
Indirect methods are inherently robust against noise, illumination change, and outliers as a result of the use of feature descriptors.
Retaining this advantage, we reconstruct detailed 3D maps by establishing dense point correspondence.
The contributions of this study are as follows: We first introduce a new dense feature point tracking algorithm based on dominant flow estimation and curvature extrema tracing.
This allows VITAMIN-E to process an enormous number of feature points; however, the need to maintain them simultaneously might lead to a high computational cost.
Therefore, we also introduce a novel optimization technique, called subspace Gauss--Newton method, for bundle adjustment.
The optimization technique significantly improves the efficiency of bundle adjustment by partially updating the variables.
Moreover, VITAMIN-E generates meshes from the reconstructed feature points and integrates them using a truncated signed distance function (TSDF)\cite{TSDF1, TSDF2, TSDF3}.
Compared with not only conventional indirect methods but also state-of-the-art direct methods, VITAMIN-E provides highly detailed 3D geometry as shown in Figure \ref{fig:FirstFig} in real time using only a CPU

\section{Dense Feature Point Tracking}

\subsection{Feature Point Tracking}

Indirect methods that use image descriptors can be unstable because of incorrect feature point correspondences.
They build feature point correspondences between multiple views by matching the descriptors.
Extracting consistent descriptors over multiple frames, however, becomes difficult because descriptors vary as the monocular camera changes its pose.
Methods such as the KLT tracker\cite{KLT} that continuously track feature points while updating the feature descriptors might be useful for overcoming the problem.
However, because the tracked positions drift as a result of a minute change of feature descriptors, the correspondences over multiple views tend to be incorrect.
These problems originate with the use of feature descriptors.

Rather than associating feature points based on descriptors, VITAMIN-E tracks the local extrema of curvature in incoming images.
In the proposed method, feature points denote the extrema of curvature on image intensities.
Let $f(x,y)$ be an image, then curvature $\kappa$ of image $f$ is as follows:
\begin{equation}
    \kappa = f_{y}^{2} f_{xx} - 2 f_{x} f_{y} f_{xy} + f_{x}^{2} f_{yy},
    \label{eq:curvature}
\end{equation}
where $f_{x}$ represents the partial derivative of $f$ with respect to $x$, which can be obtained using a Sobel operator or similar technique.
VITAMIN-E builds point correspondences over multiple images by tracking the local maximum point of curvature $\kappa(x,y,t)$, which is the extension of $\kappa$ to time domain $t$.
Figure \ref{fig:MainArt}(a) shows an example scene from which a large number of extrema of curvature $\kappa$ are extracted.
Whereas conventional indirect methods rely only on feature points with a large curvature to obtain stable correspondences, the proposed method tracks all detected extrema to reconstruct detailed geometry.

\subsection{Dominant Flow Estimation}

After detecting the extrema of curvature, the proposed method estimates a dominant flow that represents the average of optical flow over the images, which provides a good initial value to extrema tracking and makes it significantly stable, as explained later.
Specifically, we determine the corresponding feature pairs between current and previous images using the BRIEF\cite{BRIEF} feature.
Because we only have to identify coarse feature pairs over consecutive frames at this moment, feature matching is performed on low-resolution images, subsampled to 1/6 of the former size.

Then, we fit the affine transformation model $\bm{y} = A\bm{x} + \bm{b}$ to the feature pairs.
$\bm{x}$ and $\bm{y}$ denote the position of a feature point in the previous and current frame, respectively, and $A$ and \bm{$b$} represent a matrix of $2\times 2$ and a 2D translation, respectively.
$A$ and $\bm{b}$ are obtained by minimizing cost function $E$ using the Gauss--Newton method:
\begin{equation}
    E = \sum_{i}^{N} \rho\left(\|\bm{y}_i - \left(A \bm{x}_i + \bm{b}\right) \|_2\right),
    \label{eq:affine}
\end{equation}
where $N$ and $\rho$ denote the total number of corresponding points and a kernel function for M-estimation, respectively.
The following Geman--McClure kernel with scale parameter $\sigma$ is used in VITAMIN-E :
\begin{equation}
    \rho(x) = \frac{x^2}{x^2+\sigma^2}.
    \label{eq:GemanMcClure}
\end{equation}

As a result of M-estimation, the dominant flow represented by $A$ and $b$ can be estimated stably, even for low-resolution images, and it allows us to roughly predict the position of feature points in the next frame.
Note that VITAMIN-E does not rely on conventional feature matching in its core but only for prior information for dense extrema tracking, as described in the next section.
Whereas conventional feature matching has difficulty in making all feature points couple correctly between consecutive frames, affine transformation is easily obtained when at least three correspondences are given.

\subsection{Curvature Extrema Tracking}

VITAMIN-E tracks feature points by tracing the extrema of image curvature by making use of the dominant flow.
Because it depends only on extrema instead of feature descriptors used in conventional indirect methods, VITAMIN-E is free from the variation of feature descriptors caused by image noise or illumination changes, which makes VITAMIN-E highly robust.

According to the dominant flow represented by $A$ and $b$, we first predict a current position $\bar{\bm{x}}_{t_1}$ of tracking point $\bm{x}_{t_0}$:
\begin{equation}
    \bar{\bm{x}}_{t_1} = A \bm{x}_{t_0} + \bm{b}.
    \label{eq:predict}
\end{equation}
Next, prediction $\bar{\bm{x}}_{t_1}$ is corrected to ${\bm{x}}_{t_1}$ by maximizing evaluation function $F$:
\begin{equation}
    F = \kappa\left(\bm{x}_{t_1},t_1\right) + \lambda w\left(\|{ \bm{x}_{t_1} - \bar{\bm{x}} }_{t_1}\|_2\right),
    \label{eq:tracking}
\end{equation}
where $\kappa$ stores the curvature in each pixel, and $w(x)=1-\rho(x)$ and $\lambda$ denote an evaluation function and weight for the prediction, respectively.
The maximization is performed using the hill climbing method, with $\bar{\bm{x}}_{t_1}$ as the initial position.
Specifically, maximum point $\bm{x}_{t_1}$ is obtained by iterating the hill climbing method in eight neighboring pixels at each step until it reaches the local maximum value of $F$.
Function $w$ prevents the maximization process from falling into wrong extrema, thereby playing a regularization role.

Note that extrema tracking can easily fall into local solutions because there are many extrema in image curvature and it is almost impossible to distinguish them without any descriptors.
However, the prediction according to the dominant flow boosts the accuracy of extrema tracking and enables it to approach the optimal solution.

\section{Bundle Adjustment for Dense Tracking}

\subsection{Bundle Adjustment}

Bundle adjustment iteratively adjusts the reconstructed map by minimizing reprojection errors.
Given the $i$-th 3D point $\bm{p}_i$, the $j$-th camera's rotation $R_j$ and translation $\bm{t}_j$, and the 2D position $\bm{u}_{ij}$ of $\bm{p}_i$ observed in the $j$-th camera frame, the objective function is formulated as follows:
\begin{equation}
    E = \sum_{i}^{N} \sum_{j}^{M} \rho\left(\| \bm{u}_{ij} - \phi\left( R_j^T\left(\bm{p}_i-\bm{t}_j\right) \right)\|_2 \right),
    \label{eq:BACost}
\end{equation}
where $N$ and $M$ are the numbers of feature points and camera poses respectively, $\phi$ denotes the 3D-2D projection function, and $\rho$ is a kernel function for M-estimation.
Specifically, optimal camera variables $\bm{c} _j$ = ($R_j$, $\bm{t}_j$) and $\bm{p}_i$ are obtained by applying the Gauss--Newton method to Equation \ref{eq:BACost}, which results in iteratively solving the following equations:
\begin{equation}
    H \delta \bm{x} = -\bm{g}, \;\;\;\; \bm{x} = \bm{x} + \delta \bm{x},
    \label{eq:LinearEq}
\end{equation}
where $\bm{x}=(\bm{c}_1,\cdots,\bm{c}_M,\bm{p}_1,\cdots,\bm{p}_N)^T$, and $H$ and $\bm{g}$ represent the Hessian matrix and gradient around $\bm{x}$ of $E$, respectively.
$H$ and $\bm{g}$ can be represented by the camera variable $\bm{c}_j$ block and the feature point variable $\bm{p}_i$ block as follows:
\begin{equation}
    H =
    \begin{bmatrix}
        H_{cc}   & H_{cp} \\
        H_{cp}^T & H_{pp} 
    \end{bmatrix}, \;\;\;\;
    \bm{g} = 
    \begin{bmatrix}
        \bm{g}_{c} \\
        \bm{g}_{p} 
    \end{bmatrix}. 
    \label{eq:HesseGrad}
\end{equation}
Hessian matrix $H$ in bundle adjustment has a unique structure: $H_{cc}$ and $H_{pp}$ are sparse matrices with only diagonal elements in block units, whereas $H_{cp}$ is a dense matrix.
Efficient solutions that focus on this unique structure are the keys to developing highly precise and robust visual SLAM.

State-of-the-art monocular SLAM methods, such as ORB-SLAM\cite{ORB-SLAM} and DSO\cite{DSO}, solve Equation \ref{eq:LinearEq} by decomposing it using the Schur complement matrix instead of directly solving it:
\begin{eqnarray}
    \label{eq:Schur1} \left(H_{cc}-H_{cp}H_{pp}^{-1}H_{cp}^T\right) \delta \bm{x}_c &\!\!\!\!=\!\!\!\!& -\bm{g}_{c}+H_{cp}H_{pp}^{-1}\bm{g}_{p}, \\
    \label{eq:Schur2} H_{pp} \delta \bm{x}_p &\!\!\!\!=\!\!\!\!& -\bm{g}_{p} - H_{cp}^T \delta \bm{x}_c, 
\end{eqnarray}
where $\bm{x}_c=(\bm{c}_1,\cdots,\bm{c}_M)^T$ and $\bm{x}_p=(\bm{p}_1,\cdots,\bm{p}_N)^T$.
The decomposition allows us to solve bundle adjustment faster. The number of camera variables $M$ is remarkably smaller than that of feature point variables, and the inverse matrix of $H_{pp}$ can be easily calculated because it has only diagonal components; thus, the Schur complement matrix $\left(H_{cc}-H_ {cp}H_{pp}^{-1}H_{cp}^T\right)$ whose size is $M\times M$ is significantly tiny compared with the original $H$, and the inverse matrix is rapidly computable.

Equation \ref{eq:Schur1} is also called marginalization.
When regarding $H_ {cc}-H_{cp}H_{pp}^{-1}H_{cp}^T$ as a new $H$ and $\bm{g}_{c}-H_ {cp}H_{pp}^{-1}\bm{g}_{p}$ as a new $\bm{g}$, the decomposition is equivalent to eliminating all feature point variables $p$ from cost function $E$.
State-of-the-art SLAMs make themselves efficient using the marginalization technique to prevent the increase in computational cost caused by a large number of variables.

However, in the case of maintaining thousands of feature points in every frame, as in the dense extrema tracking in VITAMIN-E, the size of matrix $H$ fundamentally cannot be made sufficiently small because variable elimination is applicable only to old variables unrelated to the current frame for stability.
Moreover, the size of the Schur complement matrix is proportional to the number of feature points; thus, the calculation cost of bundle adjustment over tens of thousands points, where the size of $H$ is 100,000 $\times$ 100,000 or more, becomes too high to run bundle adjustment in real time.

\subsection{Subspace Gauss--Newton Method}

To deal with the explosion in the size of $H$, we propose a novel optimization technique called the ``subspace Gauss--Newton method.''
It partially updates variables rather than updating all of them at once, as in Equations \ref{eq:Schur1} and \ref{eq:Schur2}, by decomposing these equations further as follows:
\begin{multline}
    H_{c_i c_i} \delta \bm{x}_{c_i} = -\biggl(\bm{g}_{c_i} + 
        \sum_{l=1}^{i-1} H_{c_i c_l} \delta \bm{x}_{c_l} + \\ 
        \sum_{r=i+1}^{M} H_{c_i c_r} \delta \bm{x}_{c_r} +
    \sum_{j=1}^{N} H_{c_i p_j} \delta \bm{x}_{p_j} \biggr),
    \label{eq:GaussSeidelCam}
\end{multline}
\vspace{-5mm}
\begin{multline}
    H_{p_j p_j} \delta \bm{x}_{p_j} = -\biggl(\bm{g}_{p_j} + 
        \sum_{l=1}^{j-1} H_{p_j p_l} \delta \bm{x}_{p_l} + \\
        \sum_{r=j+1}^{N} H_{p_j p_r} \delta \bm{x}_{p_r} +
    \sum_{i=1}^{M} H_{c_i p_j}^{T} \delta \bm{x}_{c_i} \biggr).
    \label{eq:GaussSeidelPnt}
\end{multline}
Equation \ref{eq:GaussSeidelCam} updates $\delta\bm{x}_{c_i}$ of a camera variable, and Equation \ref{eq:GaussSeidelPnt} $\delta\bm{x}_{p_j}$ of a feature point variable.
$H_{c_i c_i}$, $H_{c_i p_j}$, and $H_{p_j p_j}$ are matrices of $6 \times 6$, $6 \times 3$, and $3 \times 3$, respectively.
The subspace Gauss--Newton method iteratively solves Equations \ref{eq:GaussSeidelCam} and \ref{eq:GaussSeidelPnt} until $\delta\bm{x}_{c_i}$ and $\delta\bm{x}_{p_j}$ converge, respectively.

These formulae are extensions of the Gauss--Seidel method, which is an iterative method to solve a linear system of equations, and equivalent to solving the Gauss--Newton method by fixing all variables except the variables to be optimized.
The advantage of the proposed subspace Gauss--Newton method is that it does not require a large inverse matrix, unlike Equation \ref{eq:Schur1}, but instead, only an inverse matrix of $6 \times 6$ at most.
Additionally, as in ORB-SLAM\cite{ORB-SLAM} and DSO\cite{DSO}, further speedup is possible by appropriately performing variable elimination that sets most elements of $H_{cc}$ and $H_{pp}$ to zero.
Because the proposed optimization method limits the search space in the Gauss--Newton method to its subspace, we call it the ``subspace Gauss--Newton method.''
\footnote
{
Alternating optimization, such as our method, has been used in some contexts\cite{CMAP}\cite{BA-RI}. 
See the supplementary information for details of the novel aspect of our method.
}

\section{Dense Reconstruction}

A large number of accurate 3D points are generated in real time with VITAMIN-E as a result of the dense extrema tracking and subspace Gauss--Newton method described in previous sections.
This leads not only to point cloud generation but also allows further dense geometry reconstruction that cannot be achieved by conventional indirect methods.

{\bf Meshing and Noise Removal}: We first project the 3D points onto an image and apply Delaunay triangulation to generate triangular meshes.
Then, We use NLTGV minimization proposed by Greene et al\cite{FLaME} to remove noise on the meshes.
NLTGV minimization allows us to smooth the meshes, thereby retaining local surface structures, unlike typical mesh denoising methods such as Laplacian smoothing.
Figure \ref{fig:MainArt}(c) shows example results of Delaunay triangulation and smoothing with NLTGV minimization.

{\bf Mesh Integration in TSDF}: Finally, we integrate the meshes in a TSDF to reconstruct the entire geometry of the scene.
The TSDF represents an object shape by discretizing the space into grids that store the distance from the object surface, and can merge multiple triangular meshes by storing the average value of distances from the meshes to each grid.

\begin{table*}[!t]
    \begin{center}
        \caption
        {
            Experimental results : localization success or failure [$\checkmark$ or $\times$], localization accuracy [cm], localization success rate [\%], and number of initialization retries.
        }
        \label{tbl:Exp}
        \setlength{\tabcolsep}{2.00mm}
        {\footnotesize
            \begin{tabular}{ c|p{0.5em}rl@{\hspace{1.4em}}p{0.5em}rl@{\hspace{1.4em}}p{0.5em}rl@{\hspace{1.4em}}p{0.5em}rl@{\hspace{1.4em}}}
                \toprule
                { Sequence name   } & \multicolumn{3}{c}{Our method} & \multicolumn{3}{c}{DSO\cite{DSO}} & \multicolumn{3}{c}{ORB-SLAM\cite{ORB-SLAM}}         & \multicolumn{3}{c}{LSD-SLAM\cite{LSD-SLAM}} \\
                { (no. of images) } & \multicolumn{3}{c}{}           & \multicolumn{3}{c}{}    & \multicolumn{3}{c}{w/o loop closure} & \multicolumn{3}{c}{w/o loop closure} \\
                \midrule[1pt]
                MH01 easy      & { \small$   \checkmark$} &       12.9 $\pm$ 0.5  & cm & {\small $   \checkmark$} &        6.0 $\pm$  0.8  & cm & {\small $\bm\checkmark$} & {\bf       5.2 $\pm$ 1.1  }  & cm & {\normalsize $\times$}   & {  (44.9 $\pm$ 7.2)}   & cm \\
                (3682)          &                         &      100.0 $\pm$ 0.0  & \% &                          &      100.0 $\pm$  0.0  & \% &                          & {\bf      97.7 $\pm$ 1.6  }  & \% &                     &          28.9 $\pm$ 23.6    & \% \\
                                &            & 0 $\pm$ 0\;\;\:  &                      &               & 0 $\pm$  0\;\;\: &                     &                 & {\bf 19 $\pm$ 11 }\; &                     &                 &$-$\;\;\;\;\;\;\;&                    \\
                \hline
                MH02 easy      & {\small$   \checkmark$} &        8.8 $\pm$ 0.5  & cm & {\small $   \checkmark$} &        4.2 $\pm$  0.2  & cm & {\small $\bm\checkmark$} & {\bf       4.1 $\pm$ 0.4  }  & cm & {\normalsize $\times$}   & {  (58.3 $\pm$ 6.9)}   & cm \\
                (3040)          &                         &      100.0 $\pm$ 0.0  & \% &                          &      100.0 $\pm$  0.0  & \% &                          & {\bf      92.4 $\pm$ 1.1  }  & \% &                     &          73.0 $\pm$ 1.5     & \% \\
                                &            & 0 $\pm$ 0\;\;\:  &                      &               & 0 $\pm$  0\;\;\: &                     &             & {\bf 56 $\pm$ 6 }\;\;\:  &                     &                 &$-$\;\;\;\;\;\;\;&                    \\
                \hline
                MH03 medium    & {\small$\bm\checkmark$} & {\bf  10.6 $\pm$ 1.3} & cm & {\small $   \checkmark$} &       21.1 $\pm$  0.9  & cm & {\normalsize $     \times$} & {      (4.5 $\pm$ 0.4) }  & cm & {\normalsize $\times$}   & { (266.2 $\pm$ 61.3)}  & cm \\
                (2700)          &                         & {\bf 100.0 $\pm$ 0.0} & \% &                          &      100.0 $\pm$  0.0  & \% &                             &        48.9 $\pm$ 0.8     & \% &                     &          28.4 $\pm$ 20.7    & \% \\
                                &      &{\bf 0 $\pm$ 0} \;\:    &                      &               & 0 $\pm$  0\;\;\: &                     &                   &  0 $\pm$  0\;\;\:  &                     &                 &$-$\;\;\;\;\;\;\;&                    \\
                \hline
                MH04 difficult & {\small$\bm\checkmark$} & {\bf  19.3 $\pm$ 1.6} & cm & {\small $   \checkmark$} &       20.3 $\pm$  1.0  & cm & {\small $   \checkmark$} &           33.6 $\pm$ 9.4     & cm & {\normalsize $\times$}   & { (136.4 $\pm$ 114.3)} & cm \\
                (2033)          &                         & {\bf 100.0 $\pm$ 0.0} & \% &                          &       95.7 $\pm$  0.0  & \% &                          &           95.2 $\pm$ 0.8     & \% &                     &          27.2 $\pm$ 7.0     & \% \\
                                &      &{\bf 0 $\pm$ 0} \;\:    &                      &               & 5 $\pm$  0\;\;\: &                     &                   &  6 $\pm$  1\;\;\:  &                     &                 &$-$\;\;\;\;\;\;\;&                    \\
                \hline
                MH05 difficult & {\small$   \checkmark$} &       14.7 $\pm$ 1.1  & cm & {\small $\bm\checkmark$} & {\bf  10.2 $\pm$  0.6} & cm & {\small $   \checkmark$} &           14.9 $\pm$ 4.6     & cm & {\normalsize $\times$}   & {  (27.4 $\pm$ 16.4)}  & cm \\
                (2273)          &                         &      100.0 $\pm$ 0.0  & \% &                          & {\bf  95.5 $\pm$  0.0} & \% &                          &           90.0 $\pm$ 4.0     & \% &                     &          22.7 $\pm$ 0.5     & \% \\
                                &            & 0 $\pm$ 0\;\;\:  &                      &        &{\bf 2 $\pm$ 0} \;\:    &                      &                   & 18 $\pm$  5\;\;\,  &                     &                 &$-$\;\;\;\;\;\;\;&                    \\
                \hline
                V101 easy      & {\small$   \checkmark$} &        9.7 $\pm$ 0.2  & cm & {\small $   \checkmark$} &       13.4 $\pm$  5.8  & cm & {\small $\bm\checkmark$} & {\bf       8.8 $\pm$ 0.1  }  & cm & {\normalsize $\times$}   & {  (20.0 $\pm$ 22.8)}  & cm \\
                (2911)          &                         &      100.0 $\pm$ 0.0  & \% &                          &      100.0 $\pm$  0.0  & \% &                          & {\bf      96.6 $\pm$ 0.0  }  & \% &                     &          11.6 $\pm$ 11.2    & \% \\
                                &            & 0 $\pm$ 0\;\;\:  &                      &               & 0 $\pm$  0\;\;\: &                     &             & {\bf  1 $\pm$  0 } \;\:  &                     &                 &$-$\;\;\;\;\;\;\;&                    \\
                \hline
                V102 medium    & {\small$\bm\checkmark$} & {\bf   9.3 $\pm$ 0.6} & cm & {\small $   \checkmark$} &       53.0 $\pm$  5.5  & cm & {\normalsize $     \times$} & {     (14.5 $\pm$ 11.7)}  & cm & {\normalsize $\times$}   & {  (67.0 $\pm$ 14.0)}  & cm \\
                (1710)          &                         & {\bf 100.0 $\pm$ 0.0} & \% &                          &      100.0 $\pm$  0.0  & \% &                             &        52.0 $\pm$ 3.3     & \% &                     &          15.2 $\pm$ 0.1     & \% \\
                                &      &{\bf 0 $\pm$ 0} \;\:    &                      &               & 0 $\pm$  0\;\;\: &                     &                  & 17 $\pm$  4\;\;\:   &                     &                 &$-$\;\;\;\;\;\;\;&                    \\
                \hline
                V103 difficult & {\small$\bm\checkmark$} & {\bf  11.3 $\pm$ 0.5} & cm & {\small $   \checkmark$} &       85.0 $\pm$ 36.4  & cm & {\normalsize $     \times$} & {     (37.2 $\pm$ 20.7)}  & cm & {\normalsize $\times$}   & {  (29.3 $\pm$ 2.0)}   & cm \\
                (2149)          &                         & {\bf 100.0 $\pm$ 0.0} & \% &                          &      100.0 $\pm$  0.0  & \% &                             &        65.5 $\pm$ 8.8     & \% &                     &          11.0 $\pm$ 0.1     & \% \\
                                &      &{\bf 0 $\pm$ 0} \;\:    &                      &               & 0 $\pm$  0\;\;\: &                     &                        & 56 $\pm$ 26\: &                     &                 &$-$\;\;\;\;\;\;\;&                    \\
                \hline
                V201 easy      & {\small$   \checkmark$} &        7.5 $\pm$ 0.4  & cm & {\small $   \checkmark$} &        7.6 $\pm$  0.5  & cm & {\small $\bm\checkmark$} & {\bf       6.0 $\pm$ 0.1  }  & cm & {\normalsize $\times$}   & { (131.3 $\pm$ 20.4)}  & cm \\
                (2280)          &                         &      100.0 $\pm$ 0.0  & \% &                          &      100.0 $\pm$  0.0  & \% &                          & {\bf      95.2 $\pm$ 0.0  }  & \% &                     &          74.1 $\pm$ 8.9     & \% \\
                                &            & 0 $\pm$ 0\;\;\:  &                      &               & 0 $\pm$  0\;\;\: &                     &             & {\bf  0 $\pm$  0 } \;\,  &                     &                 &$-$\;\;\;\;\;\;\;&                    \\
                \hline
                V202 medium    & {\small$\bm\checkmark$} & {\bf   8.6 $\pm$ 0.7} & cm & {\small $   \checkmark$} &       11.8 $\pm$  1.4  & cm & {\small $   \checkmark$} &           12.3 $\pm$ 2.7     & cm & {\normalsize $\times$}   & {  (42.1 $\pm$ 9.2)}   & cm \\
                (2348)          &                         & {\bf 100.0 $\pm$ 0.0} & \% &                          &      100.0 $\pm$  0.0  & \% &                          &           99.5 $\pm$ 1.2     & \% &                     &          11.3 $\pm$ 0.2     & \% \\
                                &      &{\bf 0 $\pm$ 0} \;\:    &                      &               & 0 $\pm$  0\;\;\: &                     &                   &  0 $\pm$  0\;\;\:  &                     &                 &$-$\;\;\;\;\;\;\;&                    \\
                \hline
                V203 difficult & {\small$\bm\checkmark$} & {\bf 140.0 $\pm$ 5.2} & cm & {\small $   \checkmark$} &      147.5 $\pm$  6.6  & cm & {\normalsize $     \times$} & {    (104.3 $\pm$ 64.0)}  & cm & {\normalsize $\times$}   & {  (17.7 $\pm$ 1.6)}   & cm \\
                (1922)          &                         & {\bf 100.0 $\pm$ 0.0} & \% &                          &      100.0 $\pm$  0.0  & \% &                             &        16.8 $\pm$ 15.9    & \% &                     &          11.9 $\pm$ 0.2     & \% \\
                                &      &{\bf 0 $\pm$ 0} \;\:    &                      &               & 0 $\pm$  0\;\;\: &                     &                      & 233 $\pm$ 123\; &                     &                 &$-$\;\;\;\;\;\;\;&                    \\
                \bottomrule
            \end{tabular}
        }
    \end{center}
\end{table*}

\begin{table*}[!t]
    \vspace{-4.0mm}
    \begin{center}
        \caption
        {
            Average tracking time per frame in each method.
        }
        \label{tbl:CompTime_each}
        {\footnotesize
            \begin{tabular*}{155mm}{@{\extracolsep{\fill}}c c c c}
                \toprule
                \;\;\;\;\;\;Our method & DSO & ORB-SLAM & LSD-SLAM\;\;\;\;\;\;\\
                \midrule[1pt]
                \;\;\;\;\;\;36 msec/frame & 53 msec/frame & 25 msec/frame & 30 msec/frame\;\;\;\;\;\;\\
                \bottomrule
            \end{tabular*}
        }
    \end{center}
    \vspace{-4.0mm}
\end{table*}

\begin{table*}[!t]
    \begin{center}
        \caption
        {
            Average computation time for each process of VITAMIN-E. Whereas the front-end processes ran in parallel for each frame, the back-end processes for generating the 3D mesh model were performed at a certain interval.
        }
        \label{tbl:CompTime}
        \setlength{\tabcolsep}{1.50mm}
        {\footnotesize
            \begin{tabular}{c c c | c c}
                \toprule
                \multicolumn{3}{c|}{Front-end} & \multicolumn{2}{c}{Back-end} \\
                \hline
                Feature tracking & Localization \& mapping & Meshing \& denoising & TSDF updating \& marching cubes & TSDF updating \& marching cubes\\
                                 & & & {\footnotesize(low-resolution; voxel size $\simeq$ 15 cm)} & {\footnotesize(high-resolution; voxel size $\simeq$ 2.5 cm)}\\
                \midrule[1pt]
                36 msec/frame & 25 msec/frame & 45 msec/frame & 175 msec/time & 4000 msec/time \\
                \bottomrule
            \end{tabular}
        }
    \end{center}
    \vspace{-6.0mm}
\end{table*}

\begin{figure*}[!t]
\centering
\includegraphics[width=0.93\linewidth]{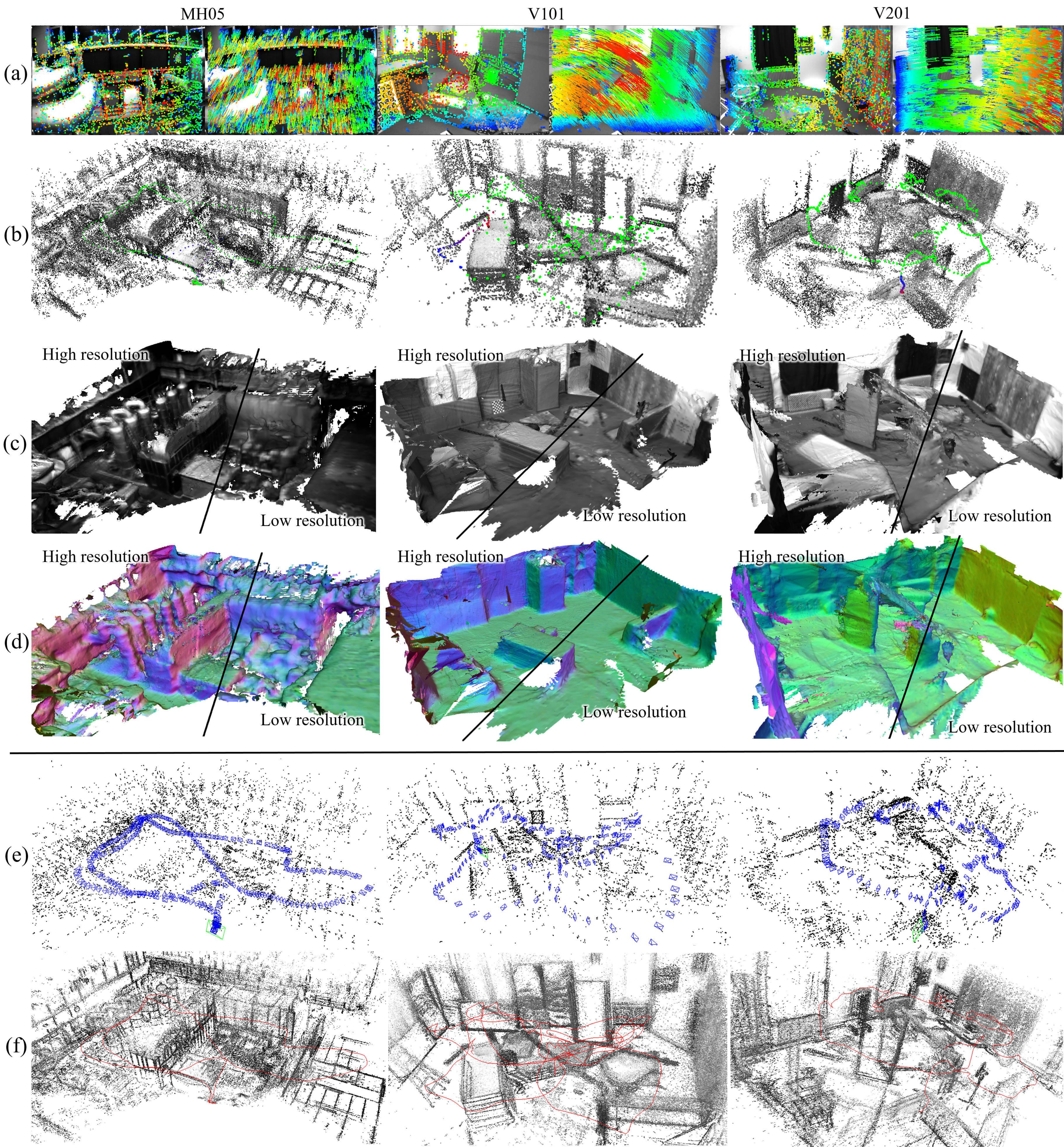}
\caption{
    Reconstruction results : (a) dense extrema tracking in real time, (b) reconstructed 3D points, (c) mesh models, and (d) normal maps generated with the proposed dense geometry reconstruction in different sized TSDF voxels, and reconstructed 3D points in the same scenes with (e) ORB-SLAM and (f) DSO.
}
\label{fig:Reconst}
\vspace{-3.0mm}
\end{figure*}

\begin{figure*}[!t]
\centering
\includegraphics[width=0.85\linewidth]{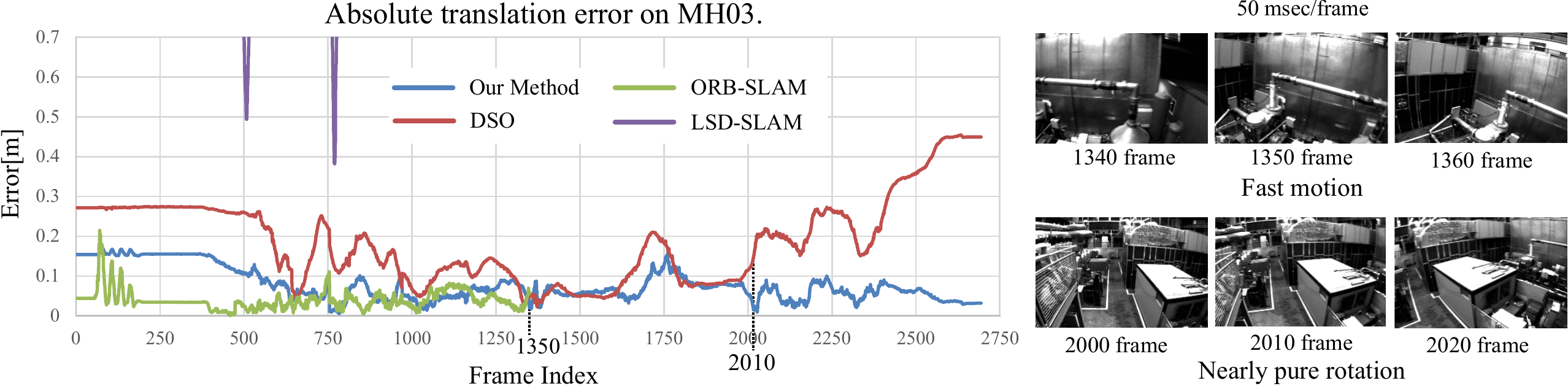}
\\
\vspace{2.0mm}
\includegraphics[width=0.85\linewidth]{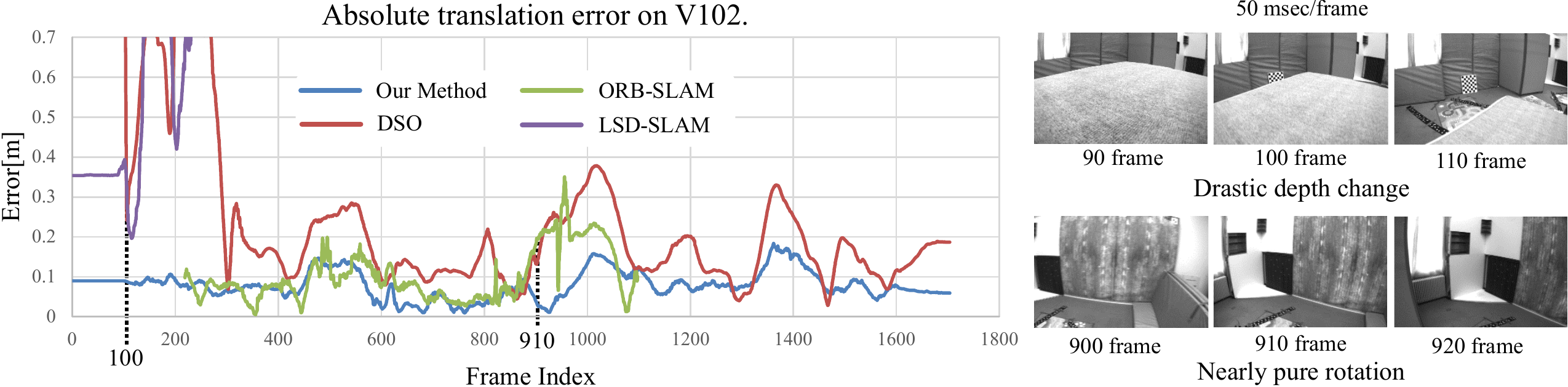}
\\
\vspace{2.0mm}
\includegraphics[width=0.85\linewidth]{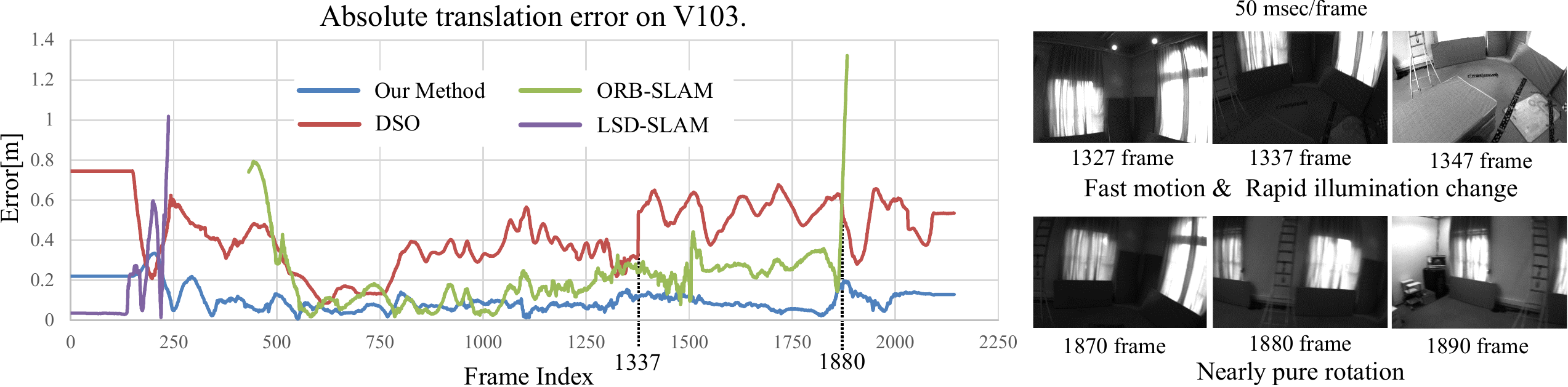}
\\
\vspace{2.0mm}
\includegraphics[width=0.85\linewidth]{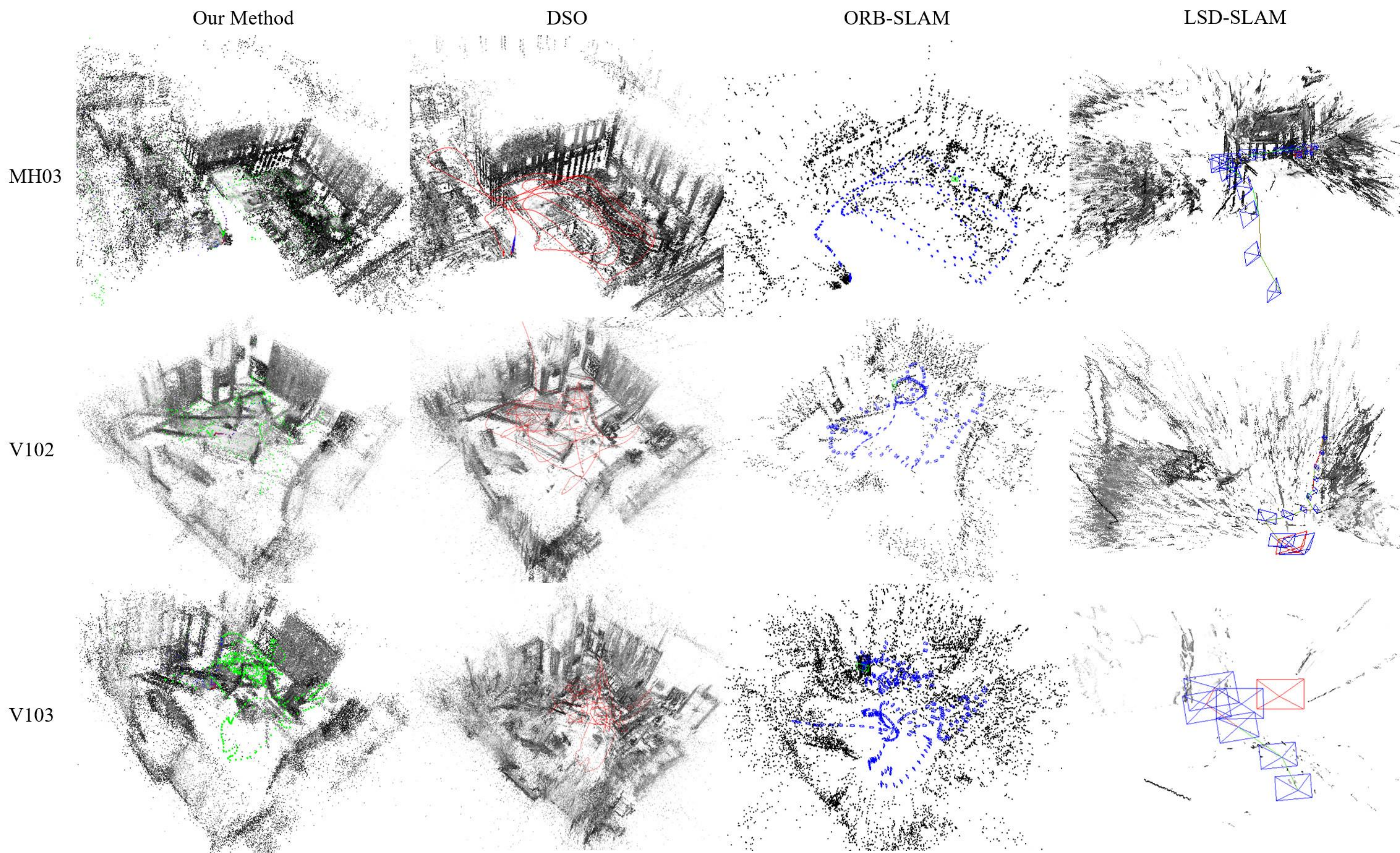}
\caption{
        Chronological RMSEs of estimated trajectories on EuRoC V102, V103, and MH03, and the reconstructed point clouds in each scene: VITAMIN-E successfully estimated the camera trajectories despite a drastic depth change, nearly pure camera rotation, rapid camera motion, and severe lighting conditions, whereas the competitors suffered from them and resulted in large trajectory errors or completely getting lost.
}
\label{fig:Graph}
\end{figure*}

\section{Experimental Results}

\subsection{Setup}

We evaluated the performance of the proposed VITAMIN-E on the visual SLAM benchmark EuRoC\cite{EuRoC}.
\footnote
{
See the supplementary information for experimental results on other datasets.
}
The dataset was created using flying drones equipped with a stereo camera and an IMU in an indoor environment, and provided ground truth of trajectories obtained by a Leica MS 50 laser tracker and Vicon motion capture.
EuRoC is also well-known for data variations, with different difficulties ranked by the movement speed and lighting conditions.
In this experiment, we compared results obtained using VITAMIN-E and other monocular SLAM methods, DSO\cite{DSO}, ORB-SLAM\cite{ORB-SLAM}, and LSD-SLAM\cite{LSD-SLAM}, using only the left images of the stereo camera in the EuRoC dataset.
Note that because VITAMIN-E and DSO\cite{DSO} do not include loop closing and relocalization, these functions in ORB-SLAM\cite{ORB-SLAM} and LSD-SLAM\cite{LSD-SLAM} were disabled to evaluate performance fairly.
Similar evaluations can be found in the papers on DSO\cite{DSO} and SVO\cite{SVO}.

VITAMIN-E ran on a Core i7-7820 HQ without any GPUs, threading each process for real time processing.
Initialization was performed using essential matrix decomposition.
Bundle adjustment is significantly sensitive to the initial value of the variable.
In VITAMIN-E, we initialized the camera variables using P3P\cite{P3P1, P3P2} with RANSAC\cite{RANSAC} and feature point variables using triangulation.
Note that the proposed bundle adjustment ran so fast that we applied it to every frame rather than each key frame as in ORB-SLAM and DSO.
To manage the TSDF, OpenChisel\cite{Chisel} was used in our implementation because of its ability to handle a TSDF on a CPU.

\subsection{Evaluation Criteria}

The evaluation on EuRoC was performed according to the following criteria:

{\bf Localization success rate}: We defined the localization success rate as $\frac{N_s}{N}$, where $N_s$ and $N$ denote the number of images that were successfully tracked and the images in the entire sequence, respectively.
If the success rate was less than 90\%, we regarded the trial as a failure.
When localization failed even once, the methods could not estimate the camera position later because loop closing and relocalization were disabled in this experiment.
Therefore, robustness greatly contributed to the success rate.

{\bf Localization accuracy}: The localization accuracy was computed by scaling the estimated trajectories so that the RMSE from ground truth trajectories was minimized because scale is not available in monocular SLAM.
Note that we did not evaluate the accuracy when the success rate was less than 90\% because the RMSE of very short trajectories tends to have an unfairly high accuracy.

{\bf Number of initialization retries}: Initialization plays an important role in monocular visual SLAM and significantly affects the success rate.
Because different methods have different initialization processes, the number of initialization retries in each method is not directly comparable, but can be a reference regarding whether the method has a weakness in initialization in certain cases.

\subsection{Results and Discussion}

Table \ref{tbl:Exp} shows the experimental results.
The results were obtained by applying each method to the image sequences in EuRoC five times, and Table \ref{tbl:Exp} shows the average values and the standard deviations of the aforementioned criteria.
Bold font is used to emphasize the highest accuracy in each sequence.
Regarding LSD-SLAM, the results of initialization retries are not included in the table because LSD-SLAM did not have a re-initialization function and failed to initialize in any sequences.
We thus manually identified the frame for which the initialization worked well in each trial, so the number of retries of LSD-SLAM was excluded from the evaluation.

The EuRoC image sequences MH01, MH02, MH04, MH05, V101, V201, and V202 are relatively easy cases for visual SLAM because the camera motion is relatively slow and the illumination does not change frequently.
By contrast, the camera moves fast in MH03, V102, V103, and V203, and additionally the lighting conditions dynamically change in V102, V103, and V203.
Furthermore, in V203, the exposure time is so long that we can see severe motion blur in the image sequence, particularly in an extremely dark environment.

Even for the challenging environments, VITAMIN-E never lost localization and outperformed other SLAM methods, DSO, ORB-SLAM, and LSD-SLAM, both in terms of accuracy and robustness, as shown in Table \ref{tbl:Exp}.
Particularly, in the sequences that contain fast camera motion, such as V102 and V103, the proposed method was superior to the existing methods.
The reconstruction results are shown in Figures \ref{fig:Reconst} and \ref{fig:Graph}.
Despite the proposed VITAMIN-E successfully generating dense and accurate geometry compared with its competitors, it performed equally fast on a CPU, as shown in Tables \ref{tbl:CompTime_each} and \ref{tbl:CompTime}.
Note that the smaller the size of the voxel in a TSDF for a detailed 3D model, the higher the computational cost.

The high accuracy and robustness of VITAMIN-E derives from tracking a large number of feature points and performing bundle adjustment for every frame.
Sharing reprojection errors among variables is important for accurate monocular SLAM, and the proposed method efficiently diffuses errors to an enormous number of variables via fast bundle adjustment.
Simultaneously, it prevents localization failure caused by losing sight of some feature points by handling a large number of feature points.

\section{Conclusion}

In this paper, we proposed a novel visual SLAM method that reconstructs dense geometry with a monocular camera.
To process a large number of feature points, we proposed curvature extrema tracking using the dominant flow between consecutive frames.
The subspace Gauss--Newton method was also introduced to maintain an enormous number of variables by partially updating them to avoid a large inverse matrix calculation in bundle adjustment.
Moreover, supported by the accurate and dense point clouds, we achieved highly dense geometry reconstruction with NLTGV minimization and TSDF integration.

VITAMIN-E is executable on a CPU in real time, and it outperformed state-of-the-art SLAM methods, DSO, ORB-SLAM, and LSD-SLAM, both in terms of accuracy and robustness on the EuRoC benchmark dataset.
Performance should be improved when loop closing is introduced in VITAMIN-E, and fusing IMU data would also be effective to stably estimate the camera pose for challenging environments such as EuRoC V203.

\noindent
{
\footnotesize
{\bf Acknowledgements:} This work was supported in part by JSPS KAKENHI, Japan Grant Numbers 16K16084 and 18K18072, and a project commissioned by the New Energy and Industrial Technology Development Organization (NEDO). 
}

{\small
\bibliographystyle{ieee_fullname}
\bibliography{egbib}
}

\clearpage
\onecolumn

\setcounter{table}{0}
\begin{table*}[b]
\begin{center}
\caption
{
Trajectory errors in the additional datasets. 
The results of other methods except our method were obtained from \cite{S_SVO}.
}%
\setlength{\tabcolsep}{3.5mm}
{\small
\begin{tabular}{l l | c c c c c c c}
\toprule
&
& 
\rotatebox[origin=l]{90}{\!\!Our method} & 
\rotatebox[origin=l]{90}{\!\!DSO} & 
\rotatebox[origin=l]{90}{\!\!SVO} & 
\rotatebox[origin=l]{90}{\begin{minipage}[t]{2.25cm}\ ORB-SLAM\\ \ {\footnotesize (with loop closure)}\end{minipage}} &
\rotatebox[origin=l]{90}{\begin{minipage}[t]{2.25cm}\ ORB-SLAM\\ \ {\footnotesize (w/o loop closure)}\end{minipage}} &
\rotatebox[origin=l]{90}{\begin{minipage}[t]{2.25cm}\ LSD-SLAM\\ \ {\footnotesize (with loop closure)}\end{minipage}} &
\rotatebox[origin=l]{90}{\begin{minipage}[t]{2.25cm}\ LSD-SLAM\\ \ {\footnotesize (w/o loop closure)}\end{minipage}}\\
\midrule[1pt]
TUM    & fr2\_desk & 1.7 cm & - & 6.7 cm & {\bf 0.9} cm & - & 4.5 cm & - \\
RGB-D  & fr2\_xyz  & 0.4 cm & - & 0.8 cm & {\bf 0.3} cm & - & 1.5 cm & - \\
\midrule[1pt]
     & Living Room 0 & {\bf  0.9} cm &      1.0  cm &  2.0 cm & - &       1.0  cm & - &      12.0  cm \\
ICL- & Living Room 1 &      11.0  cm & {\bf 2.0} cm &  7.0 cm & - &       2.0  cm & - &       5.0  cm \\
NUIM & Living Room 2 &       3.1  cm &      6.0  cm & 10.0 cm & - &       7.0  cm & - & {\bf  3.0} cm \\
     & Living Room 3 & {\bf  2.4} cm &      3.0  cm &  7.0 cm & - &       3.0  cm & - &      12.0  cm \\
\midrule[1pt]
     & Office Room 0 &      31.6  cm &     21.0  cm & 34.0 cm & - & {\bf 20.0} cm & - &      26.0  cm \\
ICL- & Office Room 1 &      40.1  cm &     83.0  cm & 28.0 cm & - &      89.0  cm & - & {\bf  8.0} cm \\
NUIM & Office Room 2 & {\bf  3.8} cm &     36.0  cm & 14.0 cm & - &      30.0  cm & - &      31.0  cm \\
     & Office Room 3 & {\bf  5.5} cm &     64.0  cm &  8.0 cm & - &      64.0  cm & - &      56.0  cm \\
\bottomrule
\end{tabular}
}
\end{center}
\vspace{-8.0mm}
\label{tbl:AddExpK}
\end{table*}

\section*{Supplementary Material}

This supplementary material provides additional experimental results and the detail of our optimization method. 

\subsection*{1.Additional Experiments}

We conducted additional experiments on the TUM RGB-D benchmark\cite{S_TUMRGBD} dataset and ICL-NUIM\cite{S_ICLNUIM} noisy synthetic dataset. 
The results are shown in Table 1, which cites scores of competitors from \cite{S_SVO} under the same conditions. The parameters of VITAMIN-E were fixed in all experiments.

Our method worked well and achieved a relatively high trajectory accuracy in a variety of scenes.
The worst case was Office Room $1$, in which the average trajectory error of VITAMIN-E was $40.1$ cm: the camera explored a textureless environment, where only edges of the floor and walls could be seen at 18 sec, which led to difficulty in extracting curvature extrema and thus unstable tracking.
Edge-based SLAMs, such as LSD-SLAM, achieved better localization performance in such scenes, whereas for point-based SLAMs' (including our method), performance tended to degrade.

\subsection*{2.Optimization}

Alternating optimization (AO) is used in some contexts such as color map optimization \cite{S_CMAP}.
In structure from motion (SfM), AO is called ``resection intersection (RI),'' which optimizes camera poses and feature positions alternately.
\cite{S_BA} concluded that although RI has numerical stability in optimization, it is not suitable for SfM because of its slow convergence compared with the standard Gauss--Newton method which optimizes all variables simultaneously.

However, we found that this is not the case in visual odometry (VO).
VO sequentially optimizes variables frame by frame; thus, most old variables are already optimized and we only need to optimize the latest camera pose and feature points.
Whereas the Gauss--Newton method optimizes all variables, the proposed subspace Gauss--Newton method inspired by RI efficiently updates partial variables.
Specifically, our method enables RI to eliminate some variables by leveraging the Schur complement.
Variable elimination plays an important role in VO because the number of variables increases with each frame.
Previous RI only solves nonlinear optimization by partially updating variables fixing the other, which results in high computational costs.
By constrast, the proposed subspace Gauss--Newton method applies RI to nonlinear optimization problems after quadratic approximation, that is, $Ax\!\!=\!\!b$, using the Schur-complement, thus enabling the real-time RI optimization required in Visual SLAM.


\end{document}